\documentclass{article}
\usepackage{ijcai20}

\usepackage{times}
\usepackage{soul}
\usepackage{url}
\usepackage[utf8]{inputenc}
\usepackage[small]{caption}
\usepackage{graphicx}
\usepackage{amsmath}
\usepackage{amsthm}
\usepackage{booktabs}
\usepackage{algorithm}
\usepackage{algorithmic}
\usepackage{todonotes}

\usepackage{xcolor}
\usepackage{amsmath}
\usepackage{mathtools}
\usepackage{multirow}

\title{Neural Abstractive Summarization with Structural Attention}

\author{
    IJCAI 2020 Submission ID 4982
}

\author{
Tanya Chowdhury$^1$
\and
Sachin Kumar$^2$ \And
Tanmoy Chakraborty$^{1}$
\affiliations
$^1$IIIT-Delhi, India\\\
$^2$Carnegie Mellon University, USA\\
\emails
\{tanya14109,tanmoy\}@iiitd.ac.in,
sachink@andrew.cmu.edu
}

\begin{document}

\maketitle

\begin{abstract}

Attentional, RNN-based encoder-decoder architectures have achieved impressive performance on abstractive summarization of news articles. However, these methods fail to account for long term dependencies within the sentences of a document. This problem is exacerbated in multi-document summarization tasks such as summarizing the popular opinion in threads present in community question answering (CQA) websites such as Yahoo! Answers and Quora. These threads contain answers which often overlap or contradict each other. In this work, we present a hierarchical encoder based on structural attention to model such inter-sentence and inter-document dependencies. We set the popular pointer-generator architecture and some of the architectures derived from it as our baselines and show that they fail to generate good summaries in a multi-document setting. We further illustrate that our proposed model achieves significant improvement over the baselines in both single and multi-document summarization settings -- in the former setting, it beats the best baseline by 1.31 and 7.8 ROUGE-1 points on CNN and CQA datasets, respectively; in the latter setting, the performance is further improved by   1.6 ROUGE-1 points on the CQA dataset.
\end{abstract}

\section{Introduction}
%Neural Abstractive summarization
Sequence-to-sequence (seq2seq) architectures with attention have led to tremendous success in many conditional language generation tasks such as machine translation \cite{seq2seqSutskever,bahdanau2014neural,luong2015effective}, dialog generation \cite{dialog}, abstractive summarization \cite{summ1,summ2,summ3} and question answering \cite{qaexample}. Attention models that compute a context vector based on the entire input text at every decoder step have especially benefited abstractive summarization. This is because -- (1) LSTM encoders struggle to encode long documents into just one vector from which one can decode the summary, and (2) generating summary involves a lot of copying from the source text which attention models easily facilitate that can also be made part of the training process \cite{summ2,nallapati2016abstractive}. 
%Ever since \cite{bahdanau2014neural} introduced an attention based seq2seq architecture for neural machine translation, architectures extending the same for summarization tasks have become very popular. A majority of these work use a bidirectional LSTM encoder and a unidirectional decoder with some formulation of attention.
% \citet{nallapati2016abstractive} in particular use a weighted attention model to calculate a context vector which is used to predict the next word in the decoder. \cite{summ2} extend their model, by adding a generation probability and introducing a  coverage loss. This allows the model to point to areas in source text when unable to generate and avoids repetition of phrases by explicitly minimizing a coverage loss. 

% The performance of these models on their chosen corpora, the CNN/Dailymail dataset has been commendable in terms of ROUGE scores. These architectures are even known to have beat the performance of (non neural) state of the art extractive summarization techniques on this dataset.
Studies in abstractive summarization majorly focus on generating news summaries, and the CNN/Dailymail dataset \cite{hermann2015teaching,nallapati2016abstractive} is largely used as a benchmark. 
% This dataset is constructed by considering the bulleted extract provided with the news article as the summary, with the rest of the news article as the input document. 
% However, the CNN/Dailymail is a NEWS dataset where most summary information can be found by just looking into topic sentences. 
% Morever, this is very clean dataset with fluent and properly structured text with a restricted domain and hardly any repetition of information. The average summary length in this dataset is limited to fifty$-$six tokens, which is not long enough for fair evaluation of a coherent summary generation task \sk{50-60 words is fairly long for single document summarization. this is not a valid argument}. Additionally, there models are trained with truncated inputs, slowly increasing to four hundred tokens \cite{summ2}. They observe that training with the full untruncated version of documents leads to a reduction in accuracy (ROUGE score). This might be because summaries information in news corpora, is often limited to topic sentences. We believe these constraints could come in the way for summarizing documents in real-world data. Here, we explore other summarization datasets and tasks, and make an attempt to propose an architecture to partially overcome these limitations.  \sk{I don't understand this, what's the motivation again? I'm rewriting this entire paragraph}
%
This is a very clean dataset with fluent and properly structured text collected from a restricted domain, and it hardly features any repetitive information. Therefore, the current models which have been built on this dataset do not need to account too much for repeated or contradicting information. Additionally, these models are trained on truncated documents (up to 400 words), and increasing the word length leads to a drop in performance of these models \cite{summ2}. This is because either  LSTM is unable to encode longer documents accurately or in this dataset,  the first few sentences contain most of the summary information. However, real-world documents are usually much longer, and their summary information is spread across the entire document rather than in the first few sentences. In this paper, we explore one such dataset dealing with community question answering (CQA) summarization.   

%Multi Document summarization and CQA introduction
Multi-document summarization (MDS) is a well-studied problem with possible applications in summarizing tweets \cite{cao2016tgsum}, news from varying sources \cite{yasunaga2017graph}, reviews in e-commerce services \cite{ganesan2010opinosis}, and so on. 
%Recent work has introduced community question answering (CQA) as an MDS task . 
%CQA services are rich knowledge resources, home to millions of archived question answer pairs. 
CQA services such as \emph{Quora}, \emph{Yahoo!~Answers}, \emph{Stack~Overflow} help in curating information which may often be difficult to obtain directly  from the existing resources on the web. However, a majority of question threads on these services harbor unstructured, often repeating or contradictory answers.
%Answers for a question thread might overlap or contradict each other. Some answers may be off topic and disjoint. In unmoderated corpora like \textit{Yahoo! Answers}, one can find the content riddled with advertisements and spam. 
Additionally, the answers vary in length from being a few words to a thousand words long. A large number of users with very different writing styles contribute to form these knowledge bases, which results in a diverse and challenging corpora. We define the task of {\em CQA summarization} as follows -- given a list of such question threads, summarize the popular opinion reflected in them. This can be viewed as a multi-document summarization task \cite{chowdhury2018cqasumm}.

We envision a solution to this problem by incorporating structural information present in language in the summarization models. Recent studies have shown promise in this direction.
%Off late, there have been attempts to study the effect of incorporating structural information in summarization. 
%However, the method for computing  structural information and its application in each work is very different. 
For example, \cite{fernandes2018structured} add graph neural networks (GNNs) \cite{li2015gated} to the seq2se  encoder model for neural abstractive summarization. \cite{liu-etal-2019-single} model documents as multi-root dependency trees, each node representing a sentence; and pick the induced tree roots to be summary sentences in an extractive setting. Recently, \cite{song2018structure} attempt to preserve important structural dependency relations obtained from syntactic parse trees of the input text in the summary by using a structure infused copy mechanism. However, obtaining labeled data to explicitly model such dependency relations is expensive. Here, we attempt to mitigate this issue by proposing a structural encoder for summarization based on the prior work on structure attention networks \cite{liu2018learning} which implicitly incorporates structural information within end-to-end training.

  Our major contributions in this work are three-fold:
  \begin{enumerate}
      \item We enhance the pointer-generator architecture by adding a structural attention based encoder to implicitly capture long term dependency relations in summarization of lengthy documents.
      \item We further propose a hierarchical encoder with multi-level structural attention to capture document-level discourse information in the multi-document summarization task.
      \item We introduce multi-level contextual attention in the structural attention setting to enable word level copying and to generate more abstractive summaries, compared to similar architectures.
  \end{enumerate}
  
  We compare our proposed solution against the popular pointer-generator model \cite{summ2} and a few other summarization models derived from it in both single and multi-document settings. We show that our structural encoder architecture beats the strong pointer-generator baseline by 1.31 ROUGE-1 points on the CNN/Dailymail dataset and by 7.8 ROUGE-1 points on the concatenated CQA dataset for single document summarization (SDS). Our hierarchical structural encoder architecture further beats the concatenated approach by another 1.6 ROUGE-1 points on the CQA dataset. A qualitative analysis of the generated summaries observe considerable qualitative gains after inclusion of structural attention.
  Our structural attention based summarinzation model is one of the few abstractive approaches to beat extractive baselines in MDS. The code is public at \url{https://bit.ly/35i7q93}.

\section{Models}
In this section, we first describe the seq2seq based pointer-generator architecture proposed by \cite{summ2,nallapati2016abstractive}. We then elaborate on how we incorporate structural attention into the mechanism and generate non-projective dependency trees to capture long term dependencies within documents. Following this, we describe our hierarchical encoder architecture and propose multi-level contextual attention to better model discourse for  MDS.

% %************************
% %BASELINE FLOWCHART
% \begin{figure}[htb]
% \begin{center}
% \includegraphics[height=2in,width=0.45\textwidth]{Images/PointerGen5.pdf}
% \caption{Poniter Generator Architecture}
% \end{center}
% \end{figure}

\subsection{Pointer-Generator Model} \label{sec:2.g}
The pointer-generator architecture (PG)  \cite{summ2} serves as the starting point for our approach. PG is inspired from the LSTM encoder-decoder architecture   \cite{nallapati2016abstractive} for the task of abstractive summarization. Tokens are fed to a bidirectional encoder to generate hidden representations $h$, and a unidirectional decoder is used to generate the summary. At the decoding step $t$, the decoder uses an attention mechanism \cite{bahdanau2014neural} to compute an attention vector $a^t$ for every encoder hidden state. These attention vectors are pooled to compute a context vector $c_t$ for every decoder step. The context vector $c_t$ at that step along with the hidden representations $h$ are passed through a feed-forward network followed by softmax to give a multinomial distribution over the vocabulary $p_\mathrm{vocab}(w)$. 
% Training is done in a teacher forcing manner with the correct input word fed at each step $t$ of the decoder. During inference the output predicted at time step $t-1$ is fed at time $t$.
% For a given decoder hidden state $s_t$, it computes attention as described by \cite{bahdanau2014neural}:
% \begin{align*}
%     e_{i}^{t} =  v^{t} tanh(W_{h}h_{i} + W_{s}s_{t} + b_{attn}) \\
%       a^{t} = softmax(e^{t}) 
% \end{align*} 

% where $v$, $W_h$, $W_s$ and $b_{attn}$ are learned during training. Further a context vector  $h^*_t$ is computed taking a weighted sum over hidden states as follows:
% \begin{equation}\label{eq:3}
    % h_t^* = \sum_i a_i^{t}h_{i}
% \end{equation}
% The context vector is then fed to multiple linear layers to generate a probability distribution over the chosen vocabulary as follows:
% \begin{equation} \label{eq:4}
%     P_{vocab} = softmax(V^{'}(V [s_t,h_t^{*}] + b)+ b^{'})   
%     \end{equation}
%  where $V^'$, $V$, $b$, $b^'$ are learned during training. This probability distribution obtained over the vocabulary is the probability distribution of the next word, i.e,   $P(w) = P_{vocab}(w)$.

Additionally, \cite{summ2} introduce a copy mechanism \cite{vinyals2015pointer} in this model. At each decoding step $t$, it predicts whether to generate a new word or copy the word based on the attention probabilities $a^t$ using a probability value $p_\mathrm{gen}$. 
% by introducing a word generation probability at each decoder time step. 
This probability is a function on the decoder input $x_t$, context vector $c_t$, and decoder state $s_t$ . 
% \begin{equation} \label{eq:7}
%     p_{gen} = \sigma (w_{h^*}^T h_t^{*} + w_{s}^{T}s_t + w_{x}^{T}x_t + b_{ptr} )
% \end{equation}
% where $w_{h^*}$,$w_{s}$,$w_{x}$ and $b_{ptr}$ are learned during training. 
$p_{gen}$ is used as a soft switch to choose between copying and generation as shown below: 
\begin{align*} \label{eq:7}
    p(w) = p_{\mathrm{gen}}p_{\mathrm{vocab}}(w) + (1- p_{\mathrm{gen}}) \sum_{i:w_i=w}a_i^t 
\end{align*}

This model is trained by minimizing  the negative log-likelihood of the predicted sequence normalized by the sequence length: $L = - \frac{1}{T} \sum_{t=0}^{T} \log p(w_t^*)$, 
% The loss function at time step $t$ and total loss for a mini batch are given by:
\if 0
\begin{align*}
    L = - \frac{1}{T} \sum_{t=0}^{T} \log p(w_t^*)
\end{align*}
\fi
where $T$ is the length of the generated summary. Once $L$ gets converged, an additional coverage loss is used to train the model further, which aims to minimize repetitions in the generated summaries. We refer the readers to \cite{summ2} for more details.
%************************
%STRUCTURE FLOWCHART

%*****************************
\subsection{Structural Attention} \label{sec:2.2}
We observe that while the pointer-generator architecture performs well on the news corpora, it fails to generate meaningful summaries on complex datasets. We hypothesize that taking into account intra-document discourse information will help in generating better summaries. We propose an architecture that implicitly incorporates such information using document structural representation  \cite{liu2018learning,kim2017structured} to include richer structural dependencies in the end-to-end training. 

\subsubsection{Proposed Model}
% Motivated by the observations by \cite{miller2016key,daniluk2017frustratingly} on the overloading performance of intermediate representations, 
% We exploit the structure of input sentences and infuse them into hidden representations by updating the semantic part of the representation with structural information.  
We model our document as a non-projective dependency parse tree by constraining inter-token attention as weights of the dependency tree. We use the Matrix tree theorem \cite{koo2007structured,tutte1984graph} to carry out the same. %For given two hidden representations at step $i$ and $j$, $h_i$ and $h_j$ respectively, We compute an inter-attention score $f_{ij} = F(h_i,h_j)$
% and normalize it to obtain $a_{ij}$ as:
% \begin{align*}
% a_{ij} = \frac{exp(f_{ij})}{\sum_{k=1}^n \exp(f_{ik})}
% \end{align*}

\begin{figure}[!t] 
\begin{center} 
\includegraphics[height=2.25in,width=0.475\textwidth]{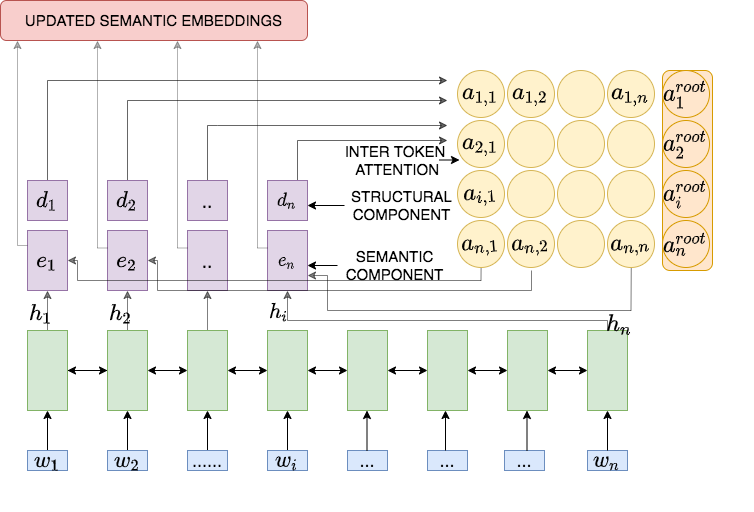}
%\vspace{-5mm}
\caption{Structural attention based encoder architecture. Tokens ($w_{i}$) are independently fed to a bi-LSTM encoder, and hidden representations $h_i$ are obtained. These representations are split into structural ($d_i$) and semantic ($e_i$) components. The structural component is used to build a non-projective dependency tree. The inter-token attention ($a_{jk}$) is computed based on  the marginal probability of an edge between two nodes  in the tree $(P(z_{jk})=1)$. %(see \Sref{sec:2.2})
}
\label{fig:1}
\end{center}
%\vspace{-5mm}
\end{figure}

As shown in Figure \ref{fig:1}, %Similar to \cite{nallapati2016abstractive,summ2}, 
we feed our input tokens ($w_i$) to a bi-LSTM encoder to obtain hidden state representations $h_i$. We decompose the hidden state vector into two parts: $d_i$ and $e_i$, which we call the structural part and the semantic part, respectively: 
\begin{align}
    [e_i,d_i] &= h_i
\end{align}
    
For every pair of two input tokens, we transform their structural parts $d$ and try to compute the probability of a parent-child relationship edge between them in the dependency tree. For tokens $j$ and $k$, this is done as:
\begin{align*}
    u_{j} = \tanh(W_{p}d_j);\ \ \ \  
    u_{k} = \tanh(W_{c}d_k)
\end{align*}
where $W_p$ and $W_c$ are learned. Next, we compute an inter-token attention function $f_{jk}$ as follows:  
\begin{align*}
    f_{jk} &= u_{k}^{T}W_{a}u_{j}
\end{align*}
 where $W_a$ is also learned. For a document with $K$ tokens, $f$ is a $K\times K$ matrix representing inter-token attention. We model each token as a node in the dependency tree and define the probability of an edge between tokens at positions $j$ and $k$, $P(z_{jk}=1)$, which is given as,
\begin{align*}
    A_{jk} &= \begin{dcases}
    0 & \text{if $j=k$}\\
    \exp(f_{jk}) & \text{otherwise}
    \end{dcases} \\
    L_{jk} &= \begin{dcases}
    \sum _{j^{'}=1}^K A_{j'k} & \text{if $j=k$}\\
    -A_{jk} & \text{otherwise}
    \end{dcases} \\
    f_j^r &= W_r d_j \\
    \bar{L}_{jk} &= \begin{dcases}
    \exp(f_j^r) & j=1 \\
    L_{jk} & j>1 \\
    \end{dcases} \\
    P(z_{jk}=1) &= (1-\delta(j,k))A_{jk}\bar{L}^{-1}_{kk} 
    -(1 -\delta(j,1))A_{jk}\bar{L}^{-1}_{kj}
\end{align*}
where $\delta(x,y) = 1$ when $x = y$ . We denote $P(z_{jk}=1)$ by $a_{jk}$ (structural attention). Let $a_{j}^{r}$ be the probability of the $j^{th}$ token to be the root: 
\begin{align}
    a_j^r &= \exp ({W_{r}d_j}) \bar{L}^{-1}_{j1}
\end{align}
We use this (soft) dependency tree formulation to compute a structural representation $r$ for each encoder token  as,
\begin{align*}
    s_i &= a_i^r e_{\mathrm{root}} + \sum_{k=1}^n a_{ki}e_k;\ \ 
    c_i = \sum_{k=1}^n a_{ik} e_i \\
    r_i &= \tanh(W_r[e_i,s_i,c_i])
\end{align*}

Thus, for encoder step $i$, we now obtain a structure infused hidden representation $r_i$. We then compute the contextual attention for each decoding time step $t$ as,
\begin{align*}
    {e_{\mathrm{struct}}}_{i}^{t} &=  v^t \tanh(W_r r_i + W_s s_t + b_{\mathrm{attn}}) \\ 
    a_{\mathrm{struct}}^{t} &= \mathrm{softmax}({e_{\mathrm{struct}}}^t)
\end{align*}

Now, using $a_{\mathrm{struct}}^{t}$, we can compute a context vector similar to standard attentional  model by weighted sum of the hidden state vectors as $
    c^{t}_{struct} = \sum_{i=1}^n a_{\mathrm{struct}_i}^t h_i
$.
At every decoder time step, we also compute the basic contextual attention vector $a^t$ (without structure incorporation), as discussed previously. We use $c^{t}_{struct}$ to compute $P_{vocab}$ and $p_{gen}$. We, however, use the initial  attention distribution $a_t$ to compute $P(w)$ in order to facilitate token level pointing.   

\begin{figure}[htb]
\begin{center} 
\includegraphics[width=\columnwidth]{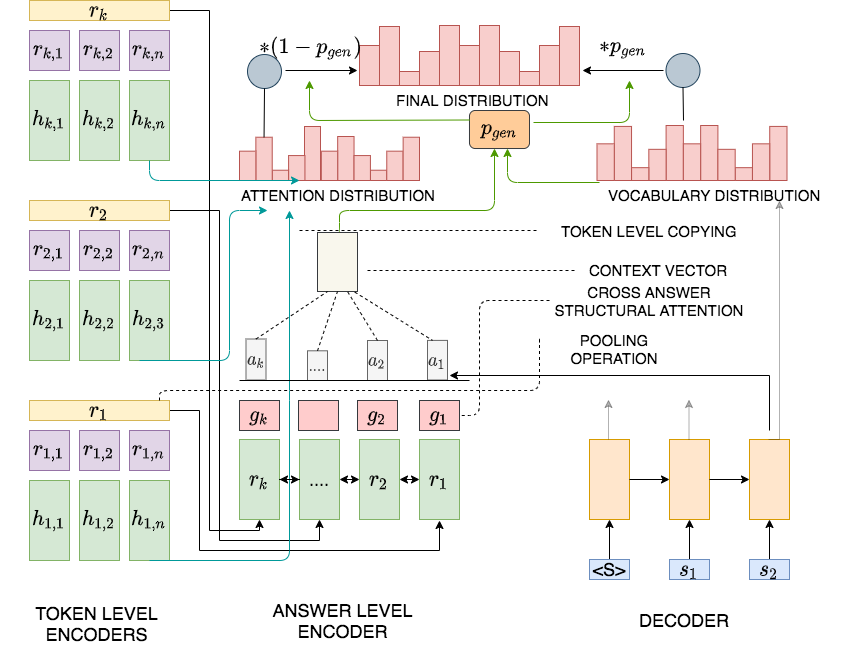}
\caption{Hierarchical encoder with structural attention + multi level contextual attention for multi-document summarization. First, token-level structured representations are computed. For each answer, these representations are pooled ($r_i$) and fed to an answer-level encoder. Structural infused  answer level embeddings ($g_i$) are calculated across answers. At every decoder time step, we calculate the context vector and probability $p_{gen}$ 
based on structural contextual attention. We calculate the copying vector (source attention) by projecting token-level attention onto the vocabulary.  }
\label{fig:2}
\end{center}
%\vspace{-5mm}
\end{figure}

\subsection{Hierarchical Model} \label{sec:2.3}

The structural attention model, while efficient, requires $\mathcal{O}(K^2)$ memory to compute attention, where $K$ is the length of the input sequence making it memory intensive for long documents. Moreover, in CQA, answers reflect different opinions of individuals. In such a case, concatenating answers results in user specific information loss. To model discourse structure better in case of conflicting (where one answer contradicts other), overlapping or varying opinion, we introduce a hierarchical encoder model based on structural attention (Figure \ref{fig:2}). 
We feed each answer independently to a bi-LSTM encoder and obtain token level hidden representations $h_{idx,tidx}$ where $idx$ is the document (or answer) index and $tidx$ is the token index. We then transform these representations into structure infused vectors $r_{idx,tidx}$ as described previously. For each answer in the CQA question thread, we pool the token representations to obtain a composite answer vector $r_{idx}$. We consider three types of pooling: \textit{average}, \textit{max} and \textit{sum}, out of which \textit{sum} pooling performs the best after initial experiments. We feed these structure infused answer embeddings to an answer-level bi-LSTM encoder and obtain higher-level encoder hidden states $h_{idx}$ from which we calculate structure infused embeddings $g_{idx}$. At the decoder time step $t$, we calculate contextual attention at answer as well as token level as follows: 
\begin{align*}
    {e_\mathrm{ans}}_{idx}^{t} &= v^{t} \tanh(W_{g}g_{idx} + W_{s}s_{t} + b_{\mathrm{attn}}) \\
    a_{\mathrm{ans}}^{t} &= \mathrm{softmax}({e_\mathrm{ans}}^{t})  \\
    {e_{\mathrm{token}}}_{idx,tidx}^{t} &= v^{t} \tanh(W_{h}h_{idx,tidx} + W_{s}s_{t} + b_{\mathrm{attn}}) \\
    {a_\mathrm{token}}^{t} &= \mathrm{softmax}({e_\mathrm{token}}^{t}) 
\end{align*}

  We use the answer-level attention distribution $a_{\mathrm{ans}}^{t}$ to compute the context vector at every decoder time step which we use to calculate $p_{vocab}$ and $p_{gen}$ as described before.
%   \begin{equation} \label{eq:25}
%       h_t^* = \sum_i {a_{ans}}i^{t}h_{i}
%   \end{equation} 
  
  To enable copying, we use $a_\mathrm{token}$. The final probability of predicting word $w$ is given by,
  %to update vocabulary scores by projecting it to an empty vocabulary vector $vocector$ as follows:
  %\begin{equation} \label{eq:26}
  %    vocvector[token\_id(a_i)] += a_i 
  %\end{equation}
%   The final distribution is computed by rewriting Eq. \ref{eq:8} as Eq. \ref{eq:27}
  \begin{align*}
    p(w) &= p_{gen} p_{vocab}(w) + (1 - p_{gen}) \sum_{i:w_i=w}{a_\mathrm{token}}_{i}^t   
  \end{align*}

\section{Datasets} \label{sec:3}
We primarily use two datasets to evaluate the performance:

(i) The {\bf CNN/Dailymail}\footnote{https://github.com/abisee/cnn-dailymail} dataset  \cite{hermann2015teaching,nallapati2016abstractive} is a news corpora containing document-summary pairs. Bulleted extracts from the CNN and Dailymail news pages online are projected as summaries to the remaining documents. The scripts released by \cite{nallapati2016abstractive} are used to extract approximately $250k$ training pairs, $13k$ validation pairs and $11.5k$ test pairs from the corpora. We use the non-anonymized form of the data to provide fair comparison to the experiments, conducted by \cite{summ2}. It is a factual, English corpora with an average document length being $781$ tokens and an average summary length being $56$ tokens. We use two versions of this dataset -- one with 400 word articles, and the other with 800 word articles. Most research reporting results on this dataset are on CNN/Dailymail-400. We also consider a 800 token version of this dataset as longer articles harbor more intra-document structural dependencies which would allow us to better demonstrate the benefits of structure incorporation. Moreover, longer documents resemble real-world datasets. (ii) We also use the {\bf CQA dataset}\footnote{https://bitbucket.org/tanya14109/cqasumm/src/master/} \cite{chowdhury2018cqasumm} which is generated by filtering the Yahoo! Answers L6 corpora to find question threads where the best answer can serve as a summary for the remaining answers. The authors use a series of heuristics to arrive at a set of $100k$ question thread-summary pairs. The summaries are generated by modifying best answers and selecting most question-relevant sentences from them. The remaining answers serve as candidate documents for summarization making up a large-scale, diverse and highly abstract dataset. On an average, the corpus has $12$ answers per question thread, with $65$ words per answer. All summaries are truncated at hundred words. We split the $100k$ dataset into $80k$ training instances, $10k$ validation and $10k$ test instances. We additionally extract the upvote information corresponding to every answer from the L6 dataset and assume that upvotes have a high correlation with the relative importance and relevance of an answer. We then rank answers in decreasing order of upvotes before concatenating them as required for several baselines. Since Yahoo! Answers is an unstructured and unmoderated question-answer repository, this has turned out to be a challenging summarization dataset \cite{chowdhury2018cqasumm}. Additionally, we  include analysis on MultiNews \cite{fabbri2019multi}, a news based MSD corpora to aid similar studies. It is the first large scale MSD news dataset consisting of 56,216 article summary pairs, crowd-sourced from various different news websites.

\section{Competing Methods}
We compare the performance of the following SDS and MDS models.
\begin{itemize}
    \item \textbf{Lead3}: It is an extractive baseline where the first 100 tokens of the document (in case of SDS datasets) and concatenated ranked documents (in case of MDS datasets) are picked to form the summary. 
    \item \textbf{KL-Summ}: It is an extractive summarization method introduced by \cite{haghighi2009exploring} that attempts to minimize KL-Divergence between candidate documents and generated summary.
    \item \textbf{LexRank}: It is an unsupervised extractive summarization method  \cite{erkan2004lexrank}. A graph is built with sentences as vertices, and edge weights are assigned based on sentence similarity. 
    \item \textbf{TextRank}: It is an unsupervised extractive summarization method which selects sentences such that the information being disseminated by the summary is as close as possible to the original documents \cite{mihalcea2004textrank}.
    \item \textbf{Pointer-Generator (PG)}: It is a supervised abstractive summarization model \cite{summ2}, as discussed earlier. It is a strong and popularly used baseline for summarization. 
    %Serves as the starting point of our proposed approach.  
    \item \textbf{Pointer-Generator + Structure Infused Copy (PG+SC)}: Our implementation is similar to one of the methods proposed by \cite{song2018structure}. We explicitly compute the dependency tree of sentences and encode a structure vector based on features like POS tag, number of incoming edges, depth of the tree, etc. We then concatenate this structural vector for every token to its hidden state representation in pointer-generator networks. 
    \item \textbf{Pointer-Generator+MMR (PG+MMR)}: It is an  abstractive MDS model \cite{lebanoff2018adapting} trained on the CNN/Dailymail dataset. It combines Maximal Marginal Relevance (MMR) method with pointer-generator networks, and shows significant performance on DUC-04 and TAC-11 datasets.
    \item \textbf{Hi-MAP}: It is an abstractive MDS model by \cite{fabbri2019multi} extending PG and  MMR.
    \item \textbf{Pointer-Generator + Structural Attention (PG+SA)}: It is the model proposed in this work for SDS and MDS tasks. We incorporate structural attention with pointer generator networks and use multi-level contextual attention to generate summaries.
    \item \textbf{Pointer-Generator + Hierarchical Structural Attention (PG+HSA)}:  We use multi-level structural attention to additionally induce a document-level non-projective dependency tree to generate more insightful summaries.
    
\end{itemize}

\begin{table}[]
    \centering
    \scalebox{0.9}{
    \begin{tabular}{l|c}
 \hline
 \textbf{Parameter} & \textbf{Value} \\
 \hline 
 Vocabulary size & 50,000 \\ 
 Input embedding dim & 128 \\
 Training decoder steps & 100 \\ 
 Learning Rate & 0.15 \\
 Optimizer & Adagrad \\
 Adagrad Init accumulator & 0.1 \\
 Max gradient norm (for clipping) & 2.0 \\
 Max decoding steps (for BS decoding) & 120 \\
 Min decoding steps (for BS decoding) &35 \\
 Beam search width & 4 \\
 Weight of coverage loss & 1 \\ 
 GPU & GeForce 2080 Ti \\
 \hline    
 \end{tabular}}
     %\vspace{-3mm}
    \caption{Parameters common to all PG-based models. }
    \label{tab:params}
    %\vspace{-4mm}
\end{table}

\begin{table}[!t]
\centering
\scalebox{0.8}{
\begin{tabular}{l|c|c|c||c|c|c}
\hline
\multirow{2}{*}{\textbf{Method}}                                        & \multicolumn{3}{c||}{\textbf{CNN/Dailymail-400}} & \multicolumn{3}{c}{\textbf{CNN/Dailymail-800}} \\ \cline{2-7}
                                                 & R-1         & R-2         & R-L        & R-1         & R-2         & R-L        \\ \hline
Lead3                                            &  40.34      &    17.70    & 36.57            &      40.34       &    17.70         &     36.57       \\
KL-Summ & 30.50 & 11.31 & 28.54 & 28.42 & 10.87 & 26.06 \\
LexRank & 34.12 & 13.31 & 31.93 & 32.36 & 11.89 & 28.12 \\
TextRank& 31.38 & 12.29 & 30.06 & 30.24 & 11.26 & 27.92 \\
PG      & 39.53 & 17.28 & 36.38 & 36.81 & 15.92 & 32.86 \\
PG+SC       &   39.82          &  17.68           &    36.72        &  37.48           &   16.61          &    33.49        \\
PG+Transformers &  39.94           &  36.44           &    36.61        &     -      &    -        &     -     \\ \hline
\textbf{PG+SA} &  \textbf{40.02}           &  \textbf{17.88}           &    36.71        &      \textbf{38.15}       &    \textbf{16.98}         &      33.20      \\ \hline
\end{tabular}}
%\vspace{-3mm}
\caption{\label{tab:ROUGE} Performance on two versions of the CNN/Dailymail dataset based on  ROUGE-1 (R-1), ROUGE-2 (R-2) and ROUGE-L (R-L) scores. }
%\vspace{-4mm}
%We truncate input documents at 400 and 800 words, respectively in the two versions of the dataset.}
\end{table}

\begin{table}[!t]
\centering
\scalebox{0.9}{
\begin{tabular}{l|c|c|c||c|c|c}
\hline
\multirow{2}{*}{\textbf{Method}}                                     & \multicolumn{3}{c||}{\textbf{CQASUMM}} & \multicolumn{3}{c}{\textbf{MultiNews}} \\ \cline{2-7}
  & \bf R-1 & \bf R-2 & \bf R-L & \bf R-1 & \bf R-2 & \bf R-L \\  \hline
Lead3  & 8.7 & 1.0  & 5.2 & 39.41 & 11.77 & 14.51 \\
KL-Summ & 24.4 & 4.3 & 13.9 & - & - & - \\
LexRank   & 28.4 & 4.7 & 14.7 & 38.27 & 12.70 & 13.20 \\
TextRank   & 27.8  & 4.8 & 14.9 & 38.40 & 13.10 & 13.50 \\ \hline
PG    & 23.2  & 4.0 & 14.6 & 41.85 & 12.91 & 16.46 \\
PG + SC & 25.5 & 4.2 & 14.8  & - & - & -\\
PG+MMR & 16.2 & 3.2 & 10 & 36.42 & 9.36 & 13.23\\
Hi-MAP & 30.9 & 4.69 & 14.9 & 43.47 & \textbf{14.87} & \textbf{17.41}\\\hline
\textbf{PG+SA} & 29.4  & 4.9 & \textbf{15.3} & 43.24 & 13.44 & 16.9 \\
\textbf{PG+HSA} &  \textbf{31.0} & \textbf{5.0} & 15.2 & \textbf{43.49} & 14.02 & 17.21 \\\hline
\end{tabular}}
%\vspace{-3mm}
\caption{Performance on the CQA   and MultiNews datasets.}\label{tab:ROUGE1}
%\vspace{-5mm}
\end{table}

\section{Experimental Results}
% We conduct both quantitative and qualitative evaluations of the models.

\subsection{Quantitative Analysis}
We compare the performance of the models on the basis of ROUGE-1, 2 and L F1-scores \cite{lin2004rouge} on the CNN/Dailymail (Table \ref{tab:ROUGE}), CQA and Multinews  (Table \ref{tab:ROUGE1}) datasets. We observe that infusion of structural information leads to considerable gains over the basic PG architecture on both the datasets. Our approach fares better than explicitly incorporating structural information \cite{song2018structure}. The effects of incorporating structure is more significant in the CQA dataset (+7.8 ROUGE-1 over PG) as compared to the CNN/Dailymail dataset (+0.49 ROUGE-1 over PG). The benefit is comparatively prominent in CNN/Dailymail-800 (+1.31 ROUGE-1 over PG) as compared to CNN-Dailymail-400 (+0.49 ROUGE-1 over PG). 

%  The pointer-generator architecture is in itself a strong baseline. In single document summarization, we notice a gain of $XX$ ROUGE-1 points compared to the pointer-generator architecture on the CNN/Dailymail dataset. We notice a massive gain of $YY$ ROUGE-1 points on the CQA dataset over our baseline. The hierarchical approach shows a further gain of $ZZ$ ROUGE-1 points over the structural attention model. We also observe that while the pointer-generator baseline is not able to beat non-neural extractive approaches on the CQA dataset, the structural approach beats the extractive baselines significantly.

\subsection{Qualitative Analysis}
We ask human evaluators\footnote{The evaluators were experts in NLP, and their age ranged between 20-30 years.}
to compare CQASUMM summaries generated by competing models on the ground of content and readability. Here we observe a significant gain with structural attention. PG has difficulty in summarizing articles with repetitive information and tends to assign lower priority to facts repeating across answers. Methods like LexRank, on the other hand, tend to mark these facts the most important. Incorporating structure solves this problem to some extent (due to the pooling operations). We find that PG sometimes picks sentences containing opposing opinions for the same summary. We find this occurrence to be less frequent in the structural attention models. This phenomenon is illustrated with an instance from the CQA dataset in Table \ref{tab:my_label}.

\subsection{Model Diagnosis and Discussion}
%Tree depth.
\paragraph{{\bf Tree depth.}} The structural attention mechanism has a tendency to induce shallow parse trees with high tree width. This leads to highly spread out trees especially at the root.

%MDS - extractive baselines are strong
\paragraph{{\bf MDS baselines. }} While supervised abstractive methods like PG  significantly outperform unsupervised non-neural methods in SDS, we find them to be inferior or similar in performance in MDS. This has also been reported by recent MDS related studies such as \cite{lebanoff2018adapting} where LexRank is shown to significantly beat basic PG on the DUC-04 and TAC-11 datasets. This justifies the choice of largely unsupervised baselines in recent MDS related studies such as \cite{yasunaga2017graph,lebanoff2018adapting}. 

%article length and dataset layout
\paragraph{{\bf Dataset layout. }} We observe that while increasing the length of the input documents lowers ROUGE score for the CNN/Dailymail dataset, it boosts the same in the CQA dataset. This might be attributed to the difference in information layout in the two datasets. The CNN/Dailymail has a very high Lead-3 score signifying summary information concentration within the first few lines of the document.

%coverage
 \paragraph{{\bf Coverage with structural attention. }} Training a few thousand iterations with coverage loss is known to significantly reduce repetition. However, in the structural attention models, while training with the CQA dataset we observe that soon after coverage loss has converged, on further training the repetition in generated summaries starts increasing again. Future work would include finding a more appropriate coverage function for datasets with repetition.  
 
% We find that on the CQA dataset, the pointer-generator baseline is unable to beat decade old extractive approaches such as LexRank \cite{erkan2004lexrank} and TextRank \cite{mihalcea2004textrank}. By observing summaries, we can conclude that the baseline architecture has trouble dealing with repetitive information. The pointer-generator approach tends to exclude repeated opinion, whereas LexRank and TextRank tend to assign them highest priority. This is why these extractive methods serve as competitive baselines for the MDS task. Our structural attention approach is successful in beating the leading baseline (TextRank) by $1.8$ points on ROUGE-1, while our hierarchical structural approach beats it by approximately $3$ points on ROUGE-1.

\section{Related Work}

\subsubsection{Neural Abstractive Summarization}
\paragraph{PG-based models.} Since the studies conducted by \cite{nallapati2016abstractive,summ2}, many approaches to neural abstractive summarization have extended them in different ways. \cite{cohan2018discourse} extend the approach proposed by \cite{summ2} with a hierarchical encoder to model discourse between various sections of a scientific paper using PubMed dataset. 
\paragraph{MDS.} Most studies on MDS were performed during DUC (2001-2007) and TAC (2008-2011) workshops. Opinosis \cite{ganesan2010opinosis} uses word-level opinion graphs to find cliques and builds them into summaries in highly redundant opinion sentences.
Recently, \cite{lebanoff2018adapting} adapt the single document architecture to a multi-document setting by using maximal marginal relevance method to select representative sentences from the input and feed to the encoder-decoder architecture. 
\cite{nema2017diversity} propose a query based abstractive summarization approach where they first encode both the query and the document, and at each decoder time step, calculate attention based on both.

\subsubsection{Structure Infused Document Modeling}
There have been numerous studies to incorporate structure in document representations by adding syntactic parsing to pipeline like architectures. However, external parser computation is costly and does not scale well.  \cite{tang2015document,tang2015learning} are one of the first to obtain document representations by first computing sentence representations and hierarchically aggregating them.  \cite{yang2016hierarchical} propose a model to implicitly add structural information within end-to-end training by assigning attention scores to each sentence according to the context at that state. \cite{kim2017structured} encoded structural information as graphical models within deep neural networks to facilitate end-to-end training. They use two graphical attention structures -- linear chain conditional random fields and graph-based parsing models, and show how both of them can be implemented as layers of a neural network. Recent work by \cite{balachandran2020structsum} proposes a similar idea by incorporating both latent and explicit sentence dependencies into single document summarization architecture. They further explicitly induce structure by injecting a coreferring
mention graph across sentences.

\begin{table}[t!]
    \centering
    \begin{small}
        \begin{tabular}{p{1.2 cm}|p{6 cm}}
        \textbf{Question} & \textbf{What came first? The chicken or the egg.?} \\ \hline
        CQA Summary & That one zygote cell divided to produce the first true chicken . Prior to that first true chicken zygote , there were only non-chickens . the zygote cell is the only place where dna mutations could produce a new animal , and the zygote cell is housed in the chicken 's egg . \textcolor{brown}{So, the egg must have as first.} \\ \hline
        Ans. 1 &  \textcolor{red}{The rooster! and he's STILL crowing about it ... always talking about this chick and that chick .... you know the type.} \\ \hline
        Ans. 2 & the chicken because it eventually evolved and layed eggs.\\ \hline
        Ans. 3 & the egg came first, well this is just my theory but i believe the egg came first because the chicken is a \textcolor{purple}{cross bred animal} or just an evolved creature of its earlier ancestors see the species before the chicken was evolving and \textcolor{olive}{eventually the species started mating and the next generation looked different so then there were chickens} im not saying im right i mean im no scientist but i like to think im pretty close \\ \hline
        Ans. 4 & \textcolor{purple}{the egg} of course...its a lot easier for an chicken egg to be a mutant of some other egg laying animal \\ \hline
        Ans. 5 & The Chicken Comes Before \textcolor{purple}{The Egg}. God Made The Chicken Before The Egg. \\ \hline
        Ans. 6 & \textcolor{orange}{An egg  containing a mutation that we now know as a chicken...} \\ \hline
        Pointer-Generator & it is a retorical example . \textcolor{teal}{egg containing changes in their dna.} in an animal like a chicken, dna from the male sperm cell and a female ovum meet and combine . that is , two \textcolor{teal}{non-chickens mated} and the dna in their new zygote contained the mutation (s) that \textcolor{teal}{produced the first true chicken .}\\ \hline
        Pointer-Generator+ Structure & an egg!! The question has not specified chicken egg! \textcolor{orange}{An egg containing a mutation we} \textcolor{blue}{today} \textcolor{orange}{no as a chicken}. It could be a dinosaur egg, dinosaur have been around for millions of years much longer than chickens!!!. \\ \hline
        Pointer-Generator + Hierarchical Structure & The egg. the chicken is a cross bred animal . its a lot easier for a chicken egg to be a mutant of some other animal. Eventually the species started mating and in the next generation, there were chickens.
    \end{tabular}
    \caption{A \textit{Yahoo!} Answers  question thread with a few of its answers. Answers 3, 4 and 6 support the `egg theory', and Answers 2, 5 support the `chicken theory'. Answer 1 is there for humor. Majority of the answers in the original document support the `egg theory'. The PG summary seems to be inconsistent, and can be seen to support both the theories within the same answer. We see that while both of our proposed models unilaterally support the `egg theory', the answer of the hierarchical model  is  framed better.}
    \label{tab:my_label}
    \end{small}
        %\vspace{-5mm}
\end{table}

% \begin{figure}
%     \centering
%     \includegraphics[height=1.8in,width=0.4\textwidth]{AAAI/Images/tree1.pdf}
%     \caption{The answer level dependency tree induced by the structural attention in the chicken-egg example for our PG+HSA model (for answers, refer to Table \ref{tab:my_label}). Answer 4 forms the root (O.A: other answers).}  
%     \label{fig:parsetree}
%     \vspace{-5mm}
% \end{figure}

\section{Conclusion}
In this work, we proposed an approach to incorporate structural attention within end-to-end training in summarization networks.
% We achieved this by computing inter-token attention by modeling them as the marginal probability of a non-projective dependency tree. We also presented a hierarchical approach to model discourse between answers in a diverse CQA dataset. 
We achieved a considerable improvement in terms of ROUGE scores compared to our primary baseline model on both the CNN/Dailymail and CQA datasets.
%A qualitative analysis indicated that structural attention improved the quality of generated summaries considerably against competitive baselines, especially on the CQA dataset. 
We also introduced multi-level contextual attention which helped in generating more abstract summaries. The analysis hinted that incorporation of some form of structural attention might be the key to achieve significant improvement compared to the extractive counterpart in the complex multi-document summarization task.

{\small 
\bibliographystyle{named}
\bibliography{ijcai20}}

\end{document}